\documentclass{svproc}
\usepackage{hyperref}

\usepackage{multirow}
\usepackage{graphicx} 
\usepackage{caption} 
\usepackage{booktabs} 

\begin{document}
\mainmatter              
\title{PointPillars Backbone Type Selection For Fast and Accurate LiDAR Object Detection}
%
\titlerunning{PointPillars Backbone Type Selection}
%
\author{Konrad Lis \href{https://orcid.org/0000-0003-2034-0590}{\includegraphics[width=16pt]{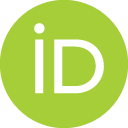}} 
\and Tomasz Kryjak \href{https://orcid.org/0000-0001-6798-4444}{\includegraphics[width=16pt]{orcid.png}}}
\authorrunning{K. Lis and T. Kryjak} 
%
%
\institute{Embedded Vision Systems Group, Computer Vision Laboratory, Department of Automatic Control and Robotics, AGH University of Science and Technology,  \\
              Al. Mickiewicza 30, 30-059 Krakow, Poland \\
              \email{kolis@agh.edu.pl, kryjak@agh.edu.pl}
}

\maketitle

\begin{abstract}
3D object detection from LiDAR sensor data is an important topic in the context of autonomous cars and drones.
In this paper, we present the results of experiments on the impact of backbone selection of a deep convolutional neural network on detection accuracy and computation speed.
We chose the PointPillars network, which is characterised by a simple architecture, high speed, and modularity that allows for easy expansion.
During the experiments, we paid particular attention to the change in detection efficiency (measured by the mAP metric) and the total number of multiply-addition operations needed to process one point cloud.
We tested 10 different convolutional neural network architectures that are widely used in image-based detection problems.
For a backbone like MobilenetV1, we obtained an almost 4x speedup at the cost of a 1.13\% decrease in mAP.
On the other hand, for CSPDarknet we got an acceleration of more than 1.5x at an increase in mAP of 0.33\%.
We have thus demonstrated that it is possible to significantly speed up a 3D object detector in LiDAR point clouds with a small decrease in detection efficiency. 
This result can be used when PointPillars or similar algorithms are implemented in embedded systems, including SoC FPGAs.
The code is available at \url{https://github.com/vision-agh/pointpillars\_backbone}.
\end{abstract}
\section{Introduction}
\label{sec:introduction}

Object detection is an important part of several systems like Advanced Driver Assistance Systems (ADAS), Autonomous Vehicles (AV) and Unmanned Aerial Vehicles (UAV).
It is also crucial for obstacle avoidance, traffic sign recognition, or object tracking (the tracking by detection approach).
Usually, object detection is related to standard vision stream processing.
Nevertheless, other sensors, such as radar, event cameras, and also LiDARs (Light Detection and Ranging) are used.
The latter has many advantages: low sensitivity to lighting conditions (including correct operation at nighttime) and a~fairly accurate 3D mapping of the environment, especially at a~short distance from the sensor.
This makes LiDAR a promising sensor for object detection, despite its high cost.
Currently, it is used in autonomous vehicles (levels 3 and 4 of the SAE classification): Waymo, Mercedes S-Class and EQS and many other experimental solutions.

\begin{figure}[!t]
    \centering
    \includegraphics[width=\textwidth]{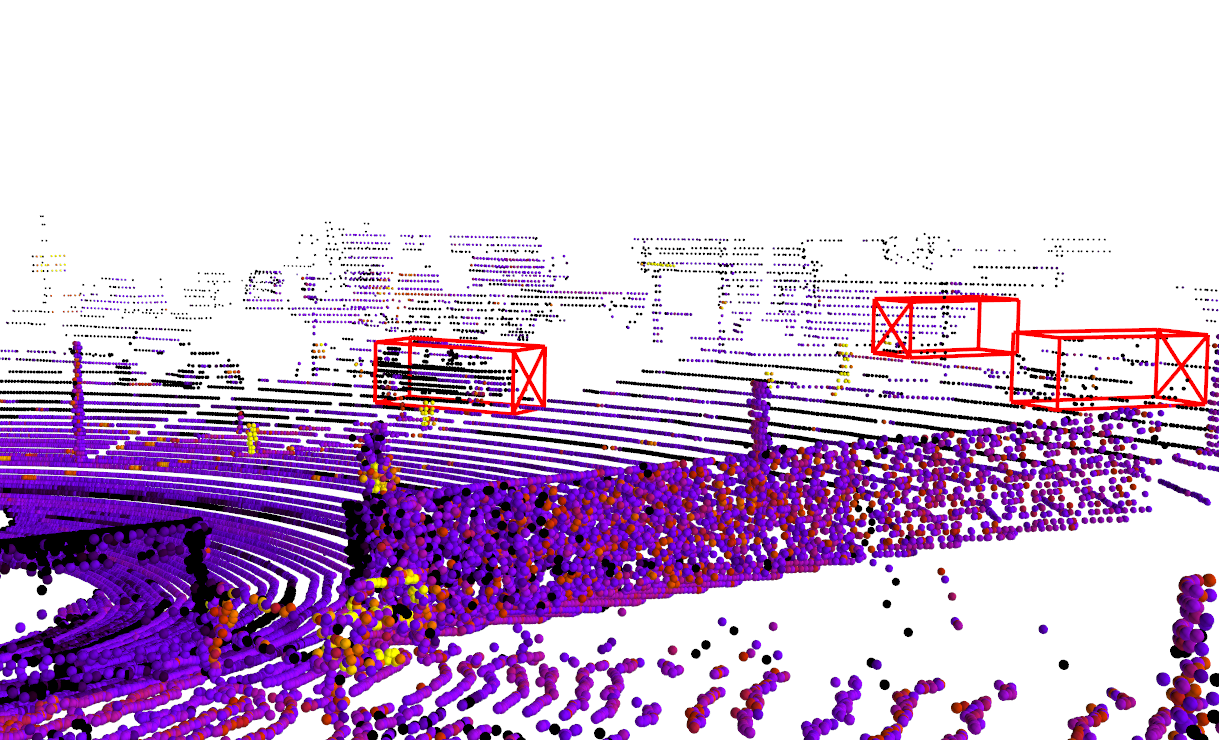}
    \caption{A~sample point cloud from the KITTI data set \cite{Kitti_www} with marked the object detections.} 
    \label{fig:cloud}
\end{figure}

It should be noted that, due to the rather specific data format, processing the 3D point clouds captured be a LiDAR significantly differs from methods known from vision systems.
The 3D point cloud is usually represented as an angle and a distance from the LiDAR sensor (polar coordinates).
Each point is also characterised by the intensity of the reflected laser beam.
Its value depends on the properties of the material from which the reflection occurred.
An example point cloud from a~LiDAR sensor with object detections is presented in Figure \ref{fig:cloud}.

In object detection systems for autonomous vehicles, the most commonly used datasets are KITTI, Wyamo Open Dataset, and NuScenes. 
KITTI Vision Benchmark Suite (2012) \cite {Kitti} is the most popular.
The training set consists of $7481$ images along with the corresponding point clouds and annotated objects.
KITTI maintains a ranking of object detection methods, where the annotated objects are split into three levels of difficulty (Easy, Moderate, Hard) corresponding to different occlusion levels, truncation, and bounding box height. 
The Waymo Open Dataset (2019) \cite{waymo} includes $1950$ sequences, which correspond to $200000$ frames, but only $1200$ sequences are annotated.
However, they contain as many as $ 12.6$ million objects.
Each year, Waymo holds few challenges in several topics, e.g. 3D object detection, motion prediction.
Nuscenes \cite{Nuscenes} contains $1000$ sequences -- it is approximately $1.4$ million images, $390$ thousand LiDAR scans, and $1.4$ million annotated objects.
NuScenes also maintains a ranking of object detection methods.
In this work, we decided to use the KITTI dataset because it still holds the position of the most widely used LiDAR database, where new solutions can be easily compared with those proposed so far.

Generally, two approaches to object detection in point clouds can be distinguished: ``classical'' and based on deep neural networks.
In the first case, the input point cloud is subjected to preprocessing (e.g. ground removal), clustering, handcrafted feature vector calculation, and classification.
``Classical'' methods achieve only moderate accuracy on widely recognised test datasets -- i.e. KITTI \cite{Kitti_www}.
In the second case, deep convolutional neural networks (DCNN) are used.
They provide excellent results (cf. the KITTI ranking \cite {Kitti_www}).
However, the price for the high accuracy is the computational and memory complexity and the need for high-performance graphics cards (Graphics Processing Units -- GPUs) -- for training and, what is even more important, inference.
This contrasts to the requirements for systems in autonomous vehicles, where detection accuracy, real-time performance, and low energy consumption are crucial.

However, these requirements can be met by selected embedded platforms.
They are characterised by low power consumption, which is usually related to lower computational power.
For real-time performance on an embedded platform, not all algorithms that run on a high-end GPU can be used.
One can choose a faster algorithm, although it usually has a negative impact on detection accuracy.
Thus, a compromise has to be made -- to speed up an algorithm for real-time performance we must accept a detection accuracy loss.
However, using several techniques like quantistion, pruning, or careful network architecture redesign, this loss can be minimised.

Embedded platforms can be divided into two rough categories: fixed or variable (reconfigurable, reprogrammable).
Examples of the former are embedded GPU devices (e.g., NVIDIA's Jetson series) and various SoC solutions that include AI coprocessors (e.g., Coral consisting of a CPU, GPU, and Google Edge TPU coprocessor).
In this case, the task is to adapt the network architecture to the platform to maximise its capabilities.
The second category of solutions are mainly SoC (System on Chip) devices containing so-called reprogrammable logic (FPGA -- Field Programmable Gate Arrays).
Examples include the Zynq SoC, Zynq UltraScale+ MPSoC and Versal/ACAP series from AMD Xilinx, and similar chips called SoC FPGAs from Intel.
Reconfigurable resources allow for greater flexibility in network implementation. 
It is possible both to adapt the network to the architecture (an example is the AMD Xilinx DPU module \cite{vai_resources_usage}) and to build custom accelerators -- then the hardware architecture is adapted to the network requirements.


Based on the initial analysis, we have selected the PointPillars \cite{pointpillars} network for experiments, mainly due to the favourable ratio of detection precision to computational complexity.
In our previous work \cite{stanisz_2021} we have described the process of running the PointPillars network on an embedded platform -- the ZCU104 evaluation board.
However, we were not able to obtain real-time performance -- a single-point cloud was processed in 374.66 ms, while it should be in less than 100ms.
Therefore, in the next step, we decided to optimise the network in terms of time performance.
The most time-consuming part of each DCNN, in terms of multiple and add (multiply-add) operations, is the so-called backbone.
In the PointPilars network, 84\% multiply-add operations are computed in the backbone.
Therefore, potentially, speeding up this part will most likely affect the overall performance of the algorithm.

In this work, we investigate the impact of replacing the PointPillars backbone with a lighter computational architecture on detection performance.
We consider 10 different backbones, inspired by fast and lightweight algorithms for object detection in images.
In particular, we are looking for solutions that significantly reduce the total number of multiply-add operations with minimal decrease in detection performance.





The main contributions of our work are:
\begin{itemize}
    \item a review of 10 different versions of PointPillars in terms of detection performance and speed,
    \item the identification of several versions of the PointPillars network that are significantly faster than the original version with only minimal decrease in detection performance.
\end{itemize}
To our best knowledge, this type of study for the PointPillars network has not been published.
We also have the shared code repository we used for the experiments.

The reminder of this paper is organised as follows.
In Section \ref{sec:related} we discuss two issues related to our work: DCNN approaches to object detection in LiDAR data and lightweight image processing DCNN backbones.
Next, in Section \ref{sec:research} we present the backbones used in our experiment.
The results obtained are summarised in Section \ref{sec:results}.
The paper ends with a short summary with conclusions and discussion of possible future work.

\section{Related work}
\label{sec:related}

LiDAR data processing has often been done with deep convolutional neural networks (DCNN).
DCNNs combine the entire processing workflow (end-to-end), including both feature extraction and classification.
They provide high recognition performance at the cost of high computational and memory complexity.
Neural networks for LiDAR data processing can be divided into two classes: 2D methods, where points from 3D space are projected onto a 2D plane under a perspective transformation, and 3D methods where no dimension is removed.
The latter are described in Section \ref{ssec:related_lidar}.
The former, given the perspective view of a~2D plane, projection-based representations are split into Front-View (FV) \cite{lasernet} and Bird’s Eye View (BEV) \cite{rt3d} representations.

\subsection{DCNN methods for 3D object detection on a LiDAR point cloud}
\label{ssec:related_lidar}

\begin{table}
\centering
\caption{Comparison of the AP results for the \textbf{3D} KITTI ranking (the \textbf{Place} column indicates the algorithm's place in the ranking). The best results are in bold. In June 2022 one of the top methods is TED. PointPillars, with an up to 11.7\% lower AP was ranked as 229th.}
\label{tab:methods_ap}
\begin{tabular}{|l|l|l|l|l|} 
\hline
\multirow{2}{*}{\textbf{Place}} & \multirow{2}{*}{\textbf{Method}} & \multicolumn{3}{c|}{\textbf{Car}}                 \\ 
\cline{3-5}
                                &                                  & \textbf{Easy}  & \textbf{Mod.}  & \textbf{Hard}   \\ 
\hline
-                               & VoxelNet                         & 77.47          & 65.11          & 57.73           \\ 
\hline
229                             & PointPillars                     & 82.58          & 74.31          & 68.99           \\ 
\hline
191                             & Patches                          & 88.67          & 77.20          & 71.82           \\ 
\hline
134                             & STD                              & 87.95          & 79.71          & 75.09           \\ 
\hline
81                              & PV-RCNN                          & 90.25          & 81.43          & 76.82           \\ 
\hline
73                              & Voxel RCNN                       & 90.90          & 81.62          & 77.06           \\ 
\hline
68                              & SIENet                           & 88.22          & 81.71          & 77.22           \\ 
\hline
24                              & SE-SSD                           & 91.49          & 82.54          & 77.15           \\ 
\hline
1                               & TED                              & \textbf{91.61} & \textbf{85.28} & \textbf{80.68}  \\
\hline
\end{tabular}
\end{table}

Based on the representations of point clouds, LiDAR-based 3D detectors can be divided into point-based \cite{point-rcnn}, voxel-based \cite{voxelnet} \cite{pointpillars} and hybrid methods \cite{PV_RCNN}. 
Point-based methods process a point cloud in an original, unstructured form.
Usually, LiDAR data are first subsampled and then processed by PointNet++ \cite{pointnet++} inspired DNNs.
An example of such a method is Point-RCNN \cite{point-rcnn}.
It first subsamples a point cloud and processes it by a~3D PointNet++ like network.
Based on point-wise features, it classifies each point to foreground or background class and generates a~bounding box proposals for foreground points.
The proposals are then filtered with Non-Maximal Suppression (NMS).
In the refinement stage, point-wise features from 3D proposals (with a small margin) are gathered and processed to obtain a final bounding box and confidence prediction.

In voxel-based methods, point clouds are first voxelized, and then a tensor of voxels is processed by 2D/3D DCNNs.
An example of such a method is PointPillars \cite{pointpillars} (described below) and VoxelNet \cite{voxelnet}.
In VoxelNet the first step is to voxelize a point cloud and apply Voxel Feature Extraction (VFE) to extract a feature vector for each voxel.
Afterward, a voxel tensor is processed by a 3D DCNN.
Then the output tensor is flattened in the Z-axis direction and fed into a 2D DCNN network to finally obtain 3D bounding boxes, class labels and confidence scores.
SECOND \cite{second} improves VoxelNet in terms of speed as it changes 3D convolutional layers in VoxelNet's backbone to sparse 3D convolutional layers.
As LiDAR data are very sparse, it significantly speeds up calculations.

Hybrid methods use elements of both aforementioned approaches.
An example is PV-RCNN \cite{PV_RCNN}.
It is a two-stage detector.
In the first stage, there are two feature extraction methods.
The first one is a SECOND-like sparse 3D DCNN which at the end generates 3D bounding box proposals.
The second is inspired by PointNet++.
First, a number of points called keypoints are sampled from a point cloud.
Then, for each stage of SECOND-like sparse 3D DCNN, for each keypoint, a set of neighbouring voxels from stage's feature map is processed by a PointNet-like network.
Outputs from these operations form a keypoint's feature vector.
In the second stage of PV-RCNN, the keypoint's feature vectors are used to refine bounding boxes.

Real-time methods focus on the speed of the algorithms which usually results in a decrease in mAP. 
Authors of \cite{r12_fast_3d} base their solution on the SECOND \cite{second} architecture.
Prior to VFE, they added a fast and efficient module which fuses LiDAR data with vision data.
The solution achieves better detection performance than SECOND and a speed of 17.8 fps (frames per second) on an Nvidia Titan RTX GPU.
The authors of \cite{r13_realtime} also took advantage of the SECOND architecture.
They calculate the initial features of the voxel as averages of the Cartesian coordinates and the intensity of the points inside the voxel -- unlike SECOND, where VFE modules are used.
They add submainfold 3D sparse convolutions to the backbone, in addition to sparse 3D convolutions.
Besides, they use self-attention mechanism and deformable convolutions.
The solution achieves 26 fps on an Nvidia RTX 2080Ti with a detection efficiency better than SECOND.
A different approach is presented by the authors of \cite{r11_pp_pruning}.
They build on the PointPillars architecture, which they accelerate using structured pruning.
They use reinforcement learning methods to determine which weights should be pruned.
They manage to get a slightly higher mAP and 76.9 fps on the Nvidia GTX 1080Ti -- a 1.5x speedup over the original PointPillars version.


The detection performance of all aforementioned algorithms is measured using Average Precision (AP), given by a formula $AP=\int_{0}^{1}p(r)dr$ where $p(r)$ is the precision in the function of recall $r$.
Usually, AP is calculated per class of the evaluation dataset.
The overall detection performance is measured with Mean Average Precision (mAP), which is AP averaged over all classes.
In Table \ref{tab:methods_ap} we present a comparison of AP results for Car detection in the KITTI dataset for several algorithms.

\subsubsection{The PointPillars}
\label{sssec:point_pillars}

\begin{figure*}
\centering
\includegraphics[width=1\textwidth]{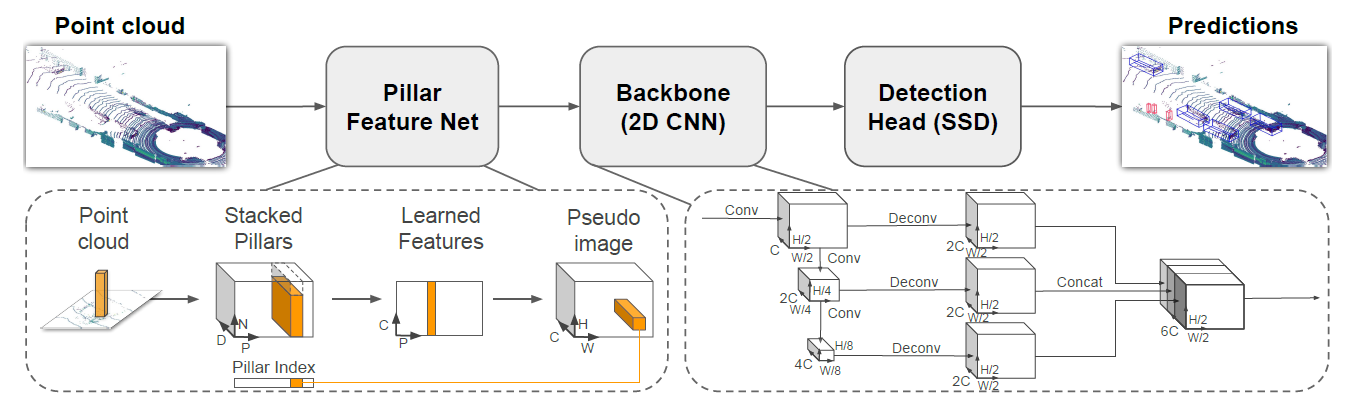}
\caption{An overview of the structure of the PointPillars network \cite{pointpillars}. The Pillar Feature Network converts the point cloud into a ``pseudo-image'', then using a 2D DCNN (with transposed convolutions), this image is transformed into a feature map used in the final detection (Single Shot Detector)}
\label{fig:pointpillars}
\end{figure*}

The PointPillars network is a voxel-based method but removes the 3D convolutions by treating the pseudo-BEV (Bird-Eye View) map as voxelized representation, so that end-to-end learning can be done using only 2D convolutions.
The input to the PointPillars \cite{pointpillars} algorithm is a~point cloud from a~LiDAR sensor.
The results are orientated cuboids that denote the detected objects.
A~``pillar'' is a~3D cell created by dividing the point cloud in the XY plane.
The network structure is shown in Figure \ref{fig:pointpillars}.
The first part, Pillar Feature Net (PFN), converts the point cloud into a sparse ``pseudo-image''.
The number of pillars is limited, as well as the number of points in each pillar.
The second part of the network -- \textit{Backbone (2D DCNN)} -- processes the ``pseudo-image'' and extracts high-level features.
It consists of two subnets: ``top-down'', which gradually reduces the dimension of the ``pseudoimage'' and another that upsamples the intermediate feature maps and combines them into the final output map.
The last part of the network is the \textit{Detection Head (SSD)}, whose task is to detect and regress the 3D cuboids surrounding the objects.
Objects are detected on a~2D grid using the Single-Shot Detector (SSD \cite{SSD}) network.
After inference, the overlapping objects are merged using the NMS algorithm.

The PointPillars network has a~rather simple architecture and relatively small computational complexity.
Moreover, it can be taken as part of other, more complex DNN like CenterPoint \cite{centerpoint}, which builds a two-stage detector based on features extracted by SECOND or PointPillars.
In the KITTI dataset, it is in the 229th\footnote{Last access: 1st June 2022} place in the 3D Car detection ranking.
For the Moderate difficulty level, it has AP equal to 74.31\% while the best method \cite{SFD} has 84.76\%.
In it's basic form PointPillars is certainly not the best available method for LiDAR based object detection.
However, nuScenes and Wyamo 3D object detection rankings contain many SOTA (State of the Art) variations of PointPillars and CenterPoint (which can incorporate PointPillars).
Due to its speed, architecture simplicity, and flexibility for enhancement, it has been chosen as a case study for our research.

\subsection{Methods for real-time object detection on images}
\label{ssec:related_image}

Real-time object detection on images can be achieved in a number of ways.
One of the most popular methods are those that form the family of YOLO single-stage detectors.
Starting from the first YOLO \cite{yolo_v1}, the authors of subsequent versions (YOLOv2 \cite{yolo_v2}, YOLOv3 \cite{yolo_v3}, YOLOv4 \cite{yolo_v4}, ScaledYOLOv4 \cite{yolo_v4_s}, YOLO-R \cite{yolo_r}, etc.) gradually improved both speed and accuracy. 
It is worth mentioning that the YOLO-R detection performance is comparable even to the best non-real-time methods.
Another example of an efficient detector is EfficientDet \cite{efficientdet} -- it is however outperformed by ScaledYOLOv4 and it's successors.

Other real-time object detection algorithms usually are formed as a general purpose fast backbone with some detection head (e.g. Single Shot Detector head).
The short history of DCNNs include several families and types of fast backbones.
NASNet \cite{nasnet} tries to use Network Architecture Search -- a reinforcement learning technique, to search for an optimal architecture for a particular task.
ResNet \cite{resnet} follows a concept of efficient skip-connected bottleneck blocks, first reducing the number of features with efficient 1x1 convolutions, then performing 3x3 convolutions, and expanding channels with subsequent 1x1 convolutions.
The authors of SqueezeNet \cite{squeezenet} had a similar idea with the \textit{Fire} module that first reduced the number of input channels with 1x1 convolutions and then performed 3x3 ones.
SqueezeNext \cite{squeezenext} tries to improve its predecessor speed, SqueezeNet, by breaking 3x3 convolutions to subsequent 1x3 and 3x1 convolutions.

There is also a whole family of Inception networks, concluded by Xception \cite{xception} and InceptionV4 \cite{inceptionv4}.
Xception, very similar to MobilenetV1 \cite{mobile_v1}, uses Depthiwse Separable Convolutions to considerably lower the number of floating point operations in the network.
Xception, in contrast to MobilenetV1, uses skip connections.
MobilenetV2 \cite{mobile_v2} improves its predecessor mainly in terms of detection accuracy.
It uses shortcut connections and introduces an Inverted Residual Layer which incorporates a 1x1 convolution layer expanding the number of channels followed by a Depthwise Separable Convolution.
MobilenetV3 \cite{mobile_v3}, on the other hand, tries to enhance its predecessor by following the NAS approach (similarly to NASNet) and proposing a few changes such as a new activation function and the usage of squeeze and excitation modules.
The authors of ShufflenetV1 \cite{shufflenetv1} use grouped and depthwise convolutions to increase speed.
To allow information flow between different groups of channels, they introduce the ``channel shuffle'' operation.
ShufflenetV2 \cite{shufflenetv2} on the other hand, splits feature map channels into two groups.
One of them is not processed at all, the other one is fed into a variant of Inverted Residual Layer (like in MobilenetV2).
After merging the groups, a ``channel shuffle'' operation is performed.

Several of the above-mentioned DNNs architectures were chosen for PointPillars backbone replacement.
Potentially, they should allow to speed up PointPillars as they speeded up networks like VGG \cite{vgg} or AlexNet \cite{alexnet}.




\section{Comparison of backbone types for the PointPillars network}
\label{sec:research}


In the original version of the PointPillars network, most of the multiply-add operations (about 84\%) are concentrated in the backbone, specifically in the ``top-down'' submodule.
Thus, potentially, speeding up this part of the algorithm will have the greatest impact on the time results obtained.
In this paper, we focus on experimenting with different computational architectures that can be used to replace the original backbone to reduce processing time at the expense of minimal decrease in detection performance.

Ten different types of architecture were selected to replace the ``top-down'' part of the backbone:
\begin{itemize}
    \item SqueezeNext \cite{squeezenext} -- consists of sequentially connected parts denoted as the ``SqueezeNext block''.
    One ``SqueezeNext block'' is a combination of a 1x1 convolution that decreases the number of channels, a 1x3 convolution, a 3x1 convolution, and a 1x1 convolution that increases the number of channels to a specified value.
    SqueezeNext in its original version consists of an input convolutional layer, 21 ``SqueezeNext blocks'' and a classifier,
    
    \item ResNet \cite{resnet} -- depending on the version, it can consist of two different types of basic blocks:
    \begin{itemize}
        \item building block -- two 3x3 convolutional layers with skip connection,
        \item bottleneck building block -- 1x1 convolution decreasing number of features, 3x3 convolution, 1x1 convolution increasing number of features and a skip connection.
    \end{itemize}
    In different ResNet versions, the total number of convolutional layers is 18, 34, 50, 101 or 152.
    
    \item ResNeXt \cite{resnext} -- the architecture is based on modified ResNet's bottleneck building block -- the 3x3 convolution is replaced by a 3x3 convolution with 32 groups.
    A grouped convolution, in contrast to an ordinary one, divides the input feature map into N groups in the channel dimension, applies the convolution operation to each of the groups separately, and concatenates the outputs in the channel dimension at the end.
    
    \item MobilenetV1 \cite{mobile_v1} -- 
    takes advantage of Depthwise Separable Convolutions (also called Separable Convolutions), a combination of ``depthwise'' and ``pointwise'' convolutions.
    A ``depthwise'' convolution, in contrast to the usual one, is performed for each channel separately and the output always has as many channels as the input.
    It is a special case of convolution with groups, where the number of groups is equal to the number of input channels.
    In MobilenetV1 the kernel size is always set to 3x3 (in ``depthwise'' convolutions).
    A ``pointwise'' convolution, on the other hand, is a 1x1 convolution, which is used to make a given output channel dependent on all input channels and possibly change the number of channels.
    The original architecture of MobilenetV1 includes a regular convolutional layer, 13 Separable Convolutions, and a classifier.
    
    \item MobilenetV2 \cite{mobile_v2} -- the basic unit is the Inverted Residual Layer.
    It consists of a 1x1 convolution, increasing the number of channels, a Separable Convolution and a skip connection (if block's stride is equal to 1).
    MobilenetV2 consists of an ordinary convolution, 7 Inverted Residual Layers and a classifier.
    
    \item ShuffleNetV1 \cite{shufflenetv1} -- the basic block is the ShuffleNet unit.
    It consists of a 1x1 group convolution, a ``channel shuffle'' operation, a ``depthwise'' 3x3 convolution, a 1x1 group convolution with the same number of groups as the first and a skip connection.
    The ``channel shuffle'' operation is designed to shuffle channels from individual convolution groups so that they can interact.
    Otherwise, using only group convolutions with the same number of groups and meanwhile ``depthwise'' convolutions, the processing paths of the individual channel groups would be completely separated from each other.
    ShufflenetV1 consists of a convolutional layer and MaxPooling, 16 ShuffleNet units and a classifier.
    
    \item ShuffleNetV2 \cite{shufflenetv2} -- the basic unit's first operation is ``channel split'' -- splits the feature map in the channel dimension into two separate maps subject to two parallel processing tracks.
    The first track leaves the map unchanged.
    The second processes the map using 3 layers: a 1x1 convolution, a ``depthwise'' 3x3 convolution, and a 1x1 convolution.
    The maps from the ends of both tracks are again combined into one by concatenation in the feature dimension.
    The final element is the ``channel shuffle'' operation, which mixes the features from both previous processing tracks.
    ShufflenetV2 consists of a 3x3 convolutional layer, MaxPooling, 16 ShuffleNet units, a 1x1 convolutional layer and a classifier.
    
    \item Darknet (Darknet53) -- the backbone of YOLOv3 \cite{yolo_v3}, the basic unit consists of a 1x1 convolutional layer, a 3x3 convolutional layer and a skip connection.
    Darknet53 has 5 blocks operating at different resolutions containing 1, 2, 8, 8, and 4 basic units respectively.
    
    \item CSPDarknet (CSPDarknet53) -- the backbone of YOLOv4 \cite{yolo_v4},
    it is based on Darknet but with different block structure.
    First, the input feature map is split in the channel dimension into two groups (similar to ShufflenetV2).
    One group is left unchanged, and the other is processed by a block of basic units from Darknet.
    Both groups are finally concatenated into one and processed by 1x1 convolution.
    
    \item Xception \cite{xception} -- it takes advantage of Separable Convolutions, the basic block consists of two or three separable convolutions with the ReLU activation function, surrounded by a skip connection.
    For blocks with stride equal to 2, after the separable convolutions a 3x3 MaxPooling with stride 2 is used and a 1x1 convolution with stride 2 is included in the skip connection to equalise the number of channels and the size of the feature map. 
    The original version of Xception consists of two convolutional layers, 12 basic blocks and a classifier.
\end{itemize}

\begin{table}[t!]
\centering
\caption{Backbones characterisation. Par. denotes number of parameters (unit:  $10^{6}$ parameters). MAdd denotes number of multiply-add operations (unit: $10^{9}$ operations). fps-B denotes number of frames per second measured for backbone only. fps denotes processing rate for the whole algorithm. MAdd-Su, fps-B-Su and fps-Su denote speedup in terms of number of multiply-add operations, in terms of backbone fps and in terms of the whole algorithm processing rate. All speedup values are calculated relative to the \textit{base} backbone version. fps and fps-B values are measured using Nvidia RTX 2070S GPU (using the mmdetection3d environment). The number of parameters is 1.5x-5.5x smaller than in the original PointPillars. There is no strict correlation between fps-B and MAdd values. Interestingly, the speedup measured in terms of fps-B and fps is significantly smaller than speedup measured in terms of number of multiply-add operations.}
\label{tab:results_speed}
\begin{tabular}{l|l|c|l|l|l|l|l} 
\toprule
                      & \textbf{Par.} & \textbf{MAdd}    & \textbf{MAdd-Su} & \textbf{fps-B} & \textbf{fps-B-Su} & \textbf{fps}  & \textbf{fps-Su}  \\ 
\cline{2-8}
\textbf{base}         & 4.83          & 34.91            & 1                & 128            & 1                 & 47.8          & 1                \\
\textbf{CSPDarknet}   & 2.33          & 20.10            & 1.74             & 127.9          & 1                 & 46.2          & 0.97             \\
\textbf{Darknet}      & 3.15          & 23.51            & 1.48             & 131.1          & 1.02              & 48.7          & 1.02             \\
\textbf{MobilenetV1}  & 1.13          & ~ 8.84           & 3.95             & 194.9          & 1.52              & 53            & 1.11             \\
\textbf{MobilenetV2}  & 1.13          & ~ 8.84           & 3.95             & 191.6          & 1.5               & \textbf{54.3} & \textbf{1.14}    \\
\textbf{ResNet}       & 1.49          & 12.28            & 2.84             & 173.3          & 1.35              & 51.7          & 1.08             \\
\textbf{ResNeXt}      & 1.49          & 12.30            & 2.84             & 148.4          & 1.16              & 48.7          & 1.02             \\
\textbf{ShufflenetV1} & 1.11          & ~ 9.78           & 3.57             & 161.3          & 1.26              & 50.9          & 1.06             \\
\textbf{ShufflenetV2} & \textbf{0.88} & ~ \textbf{7.80 } & \textbf{4.48}    & \textbf{200}   & \textbf{1.56}     & 54            & 1.13             \\
\textbf{SqueezeNext}  & 1.84          & 14.66            & 2.38             & 86.7           & 0.68              & 38.7          & 0.81             \\
\textbf{Xception}     & 1.17          & 10.72            & 3.26             & 131.6          & 1.03              & 47.1          & 0.99             \\
\bottomrule
\end{tabular}
\end{table}


Our choice was guided by both the speed of the individual networks and the relatively low complexity of the computational architecture.
For implementation on embedded systems, e.g., SoC FPGAs, the complexity and irregularity of the computing architecture can make implementation much more difficult and slower.
For this reason, we have omitted e.g. NASNet.
The main novelty of MobilenetV3 compared to MobilenetV2 is the use of the NAS (Network Architecture Search) technique and a related Netadapt technique so as to tailor the architecture for a specific computational task.
In this paper, we study other computational task than MobilenetV3 was adapted to, and the network architectures are modified anyway.
For this reason, MobilenetV3 is omitted and MobilenetV2 is examined.
We have also omitted the Inception family \cite{inceptionv4}, due to comparable results with the architecturally simpler ResNet.

Additionally, each of the architectures considered had to be modified so that it could be applied to the PointPillars network.
We assumed that the numbers of layers, the numbers of channels, and the number of blocks in the original version of PointPillars are at least roughly matched to the task of 3D object detection in the KITTI dataset.
Thus, we decided on the following solution: each Conv-BN-ReLU layer sequence from the ``top-down'' part of the original backbone was considered as a so-called ``basic unit''.
Then, for each of the considered computational architectures, a backbone of type X was created by replacing the basic units from the ``top-down'' part of the original backbone by the basic units from X.
The basic units in different backbone types are considered as:
\begin{itemize}
    \item SqueezeNext -- SqueezeNext block,
    \item ResNet -- bottleneck building block,
    \item ResNeXt -- bottleneck building block with ResNeXt modifications,
    \item MobilenetV1 -- Depthwise Separable Convolution,
    \item MobilenetV2 -- Inverted Residual Layer,
    \item ShuffleNetV1 -- ShuffleNet unit,
    \item ShuffleNetV2 -- basic unit of ShufflenetV2,
    \item Darknet -- Darknet's basic unit,
    \item CSPDarknet -- Darknet's basic unit, however, as in CSPDarknet, blocks of basic units in the modified PointPillars backbone are split into two parallel processing lanes,
    \item Xception -- Xception's basic block.
\end{itemize}
In future work, once a particular backbone is selected, NAS techniques can be used to find the optimal number of layers, number of channels, and number of blocks.
Currently, such experiments with 10 different networks would be too time-consuming.


The framework \emph{mmdetection3d} \cite{mmdet3d2020}, based on PyTorch, was chosen for the experiments.
It contains implementations of the chosen 3D object detection methods, including PointPillars.
In comparison to the original PointPillars implementation, it has better modularity and custom changes can be made more conveniently.
These two features made the experiments easier to conduct.
\emph{mmdetection3d} is less optimised in terms of speed (for PointPillars, it offers about 40 fps instead of 60 fps, voxelization and NMS are particularly slow), but this does not affect the results, since the number of multiply-add operations is compared, not the number of fps obtained on a specific hardware platform.
In terms of detection efficiency, these two implementations are comparable, since the mAP values of car detection on the KITTI-val set in \emph{mmdetection3d} (77.1\%) and the original implementation (76.9\%) are very close.
However, it should be noted that in this work the PointPillars network is trained on all three classes at the same time, as opposed to the paper \cite{pointpillars} where the network for car detection and the network for pedestrian and cyclist detection are trained separately.
As a consequence, the results may be slightly worse than those obtained in \cite{pointpillars}.


All experiments were run with the same training parameters as the base PointPillars implementation from \emph{mmdetection3d}, which in turn uses the same settings as \cite{pointpillars} except for:
\begin{itemize}
    \item optimiser -- AdamW instead of Adam,
    \item weight decay equal 0.01 -- original PointPillars do not use weight decay,
    \item learning rate schedule -- cosine annealing learning rate schedule with initial learning rate $10^{-3}$, rising to $10^{-2}$ after 64 epochs and then falling to $10^{-7}$ at the final 160th epoch -- instead of original exponentially decaying learning rate.
\end{itemize}
We used Nvidia RTX 2070S GPU, one training lasted for 14 hours on average.

\section{Results}
\label{sec:results}
    

\begin{figure*}[t]
\centering
\includegraphics[width=0.75\textwidth]{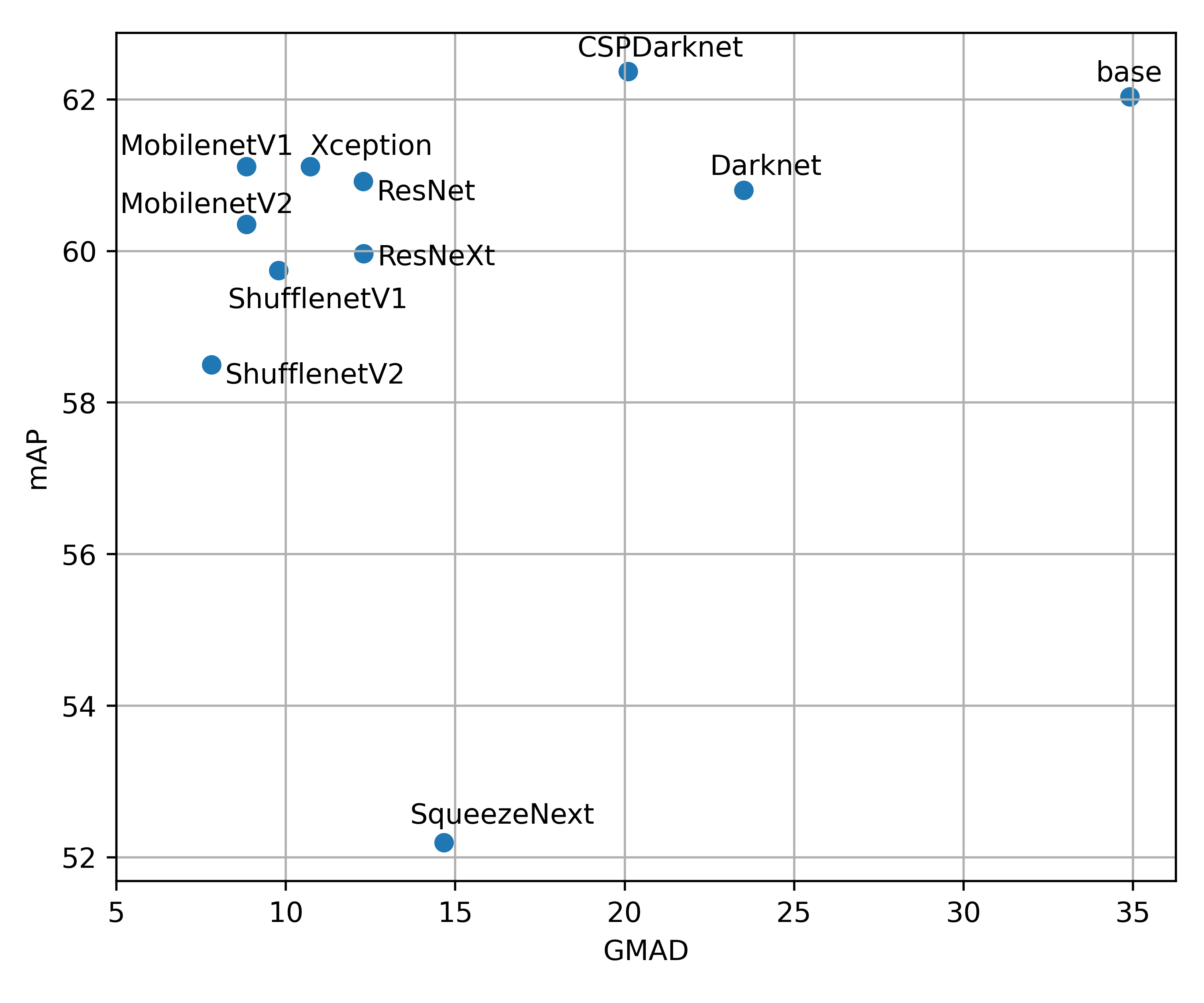}
\caption{mAP results (for all classes) compared to the number of multiply-add operations for various backbone types. GMAD unit denotes $10^{9}$ of multiply-add operations. The Pareto set includes ShufflenetV2, MobilenetV1 and CSPDarknet.}
\label{fig:map_vs_mad_overall}
\end{figure*}

Detection efficiency is measured with the mAP (Mean Average Precision) metric, which is the average AP value for all classes (Car, Pedestrian, Cyclist) and all difficulty levels (Easy, Moderate, Hard).
For network speed comparison, we use the number of multiply-add operations (MAD) needed to process one point cloud.
This value is independent of the computing platform and the computation's precision -- whether floating point, fixed point, or binary values are used.
The frame rate is not used directly for the speed comparison because, depending on the specific hardware platform and optimizations in the computing libraries, the results can vary significantly.
In the results, the original PointPillars network is denoted as ``base''.
We adopt the real-time definition from \cite{stanisz_2021}, i.e. processing point clouds at a rate of 10 fps or greater.



\begin{figure*}[t!]
\centering
\includegraphics[width=0.75\textwidth]{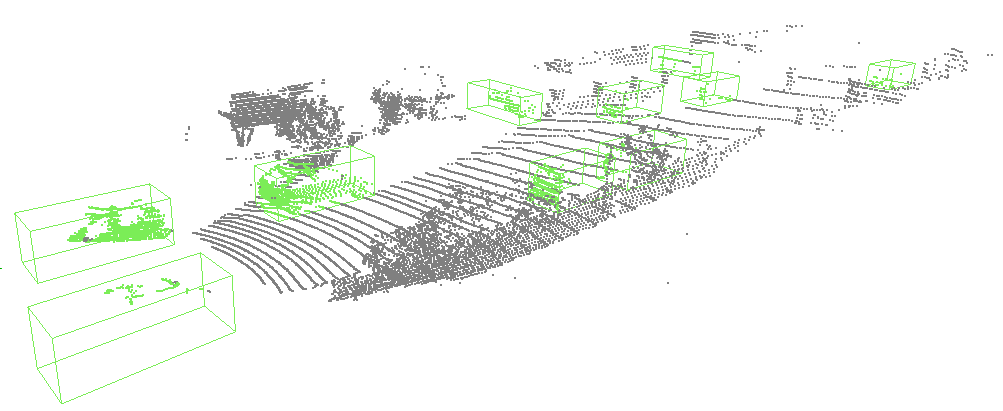}
\caption{An example detection result of PointPillars version with CSPDarknet backbone.}
\label{fig:example_detection}
\end{figure*}

The results for all KITTI classes and difficulty levels are shown in Table \ref{tab:results_ap}, an exemplary detection is shown on Figure \ref{fig:example_detection}.
It is worth noting that the original PointPillars version is not always the best.
Table \ref{tab:results_map} shows the results in a more compact form -- mAP versus the speed of each backbone type (in terms of the number of multiply-add operations).
Table \ref{tab:results_speed} shows more detailed backbones size and speed characteristics, including number of multiply-add operations, number of parameters and fps measured with Nvidia RTX 2070S GPU.
The most surprising backbone type is CSPDarknet, which has both better mAP than the original PointPillars version and more than 1.5x less multiply-add operations.


\begin{table}
\centering
\caption{AP results of 3D detection for various backbone types. Easy, Moderate (Mod.) and Hard are the KITTI object detection difficulty levels. Note that the original PointPillars does not always achieve the best AP results.}
\label{tab:results_ap}
\begin{tabular}{l|lll|lll|lll} 
\toprule
                      & \multicolumn{3}{c|}{\textbf{Car }}               & \multicolumn{3}{c|}{\textbf{Pedestrian }}        & \multicolumn{3}{c}{\textbf{Cyclist }}             \\ 
\cline{2-10}
                      & Easy           & Mod.           & Hard           & Easy           & Mod.           & Hard           & Easy           & Mod.           & Hard            \\ 
\cline{2-10}
\textbf{base}         & \textbf{85.90} & 73.88          & 67.98          & 50.17          & \textbf{45.11} & \textbf{41.09} & 78.66          & 59.51          & 56.02           \\
\textbf{CSPDarknet}   & 85.87          & \textbf{75.81} & \textbf{68.44} & \textbf{50.31} & 44.97          & 39.67          & 79.41          & 59.68          & \textbf{57.17}  \\
\textbf{Darknet}      & 84.94          & 75.49          & 68.42          & 45.85          & 40.99          & 36.53          & \textbf{79.78} & 59.45          & 55.78           \\
\textbf{MobilenetV1}  & 82.95          & 73.15          & 67.8           & 49.35          & 45.08          & 41.47          & 76.76          & 58.50          & 54.99           \\
\textbf{MobilenetV2}  & 83.85          & 73.15          & 67.88          & 48.45          & 43.76          & 39.12          & 75.69          & 57.36          & 53.88           \\
\textbf{ResNet}       & 84.61          & 73.47          & 68.03          & 47.58          & 42.70          & 38.53          & 78.50          & 59.06          & 55.80           \\
\textbf{ResNeXt}      & 84.60          & 73.86          & 68.13          & 47.11          & 42.37          & 37.73          & 75.80          & 57.07          & 52.99           \\
\textbf{ShufflenetV1} & 83.61          & 73.33          & 67.58          & 47.49          & 43.03          & 38.41          & 74.34          & 57.16          & 53.72           \\
\textbf{ShufflenetV2} & 82.85          & 72.62          & 67.42          & 43.41          & 38.82          & 35.22          & 74.68          & 57.47          & 54.03           \\
\textbf{SqueezeNext}  & 77.05          & 64.87          & 58.5           & 43.47          & 38.83          & 35.72          & 62.68          & 45.49          & 43.11           \\
\textbf{Xception}     & 84.81          & 75.37          & 68.21          & 46.12          & 41.43          & 37.13          & 78.97          & \textbf{61.20} & 56.81           \\
\bottomrule
\end{tabular}
\end{table}

\begin{table}
\centering
\caption{mAP results of 3D detection for various backbone types. mAP for Car, Pedestrian and Cyclist is AP averaged over KITTI difficulty levels for particular classes. mAP for Overall is the AP averaged across all classes and all KITTI difficulty levels. It is worth noting that the original PointPillars is not the best in Overall mAP. The Speedup is computed as a ratio of given backbone type's to original PointPillars multiply-add operations number.}
\label{tab:results_map}
\begin{tabular}{l|c|l|c|c||c|c} 
\toprule
\begin{tabular}[c]{@{}l@{}}\\\end{tabular} & \textbf{Overall} & \multicolumn{1}{c|}{\textbf{Car}} & \textbf{Pedestrian} & \textbf{Cyclist} & \textbf{GMADs}   & \multicolumn{1}{l}{\textbf{Speedup}}  \\ 
\cline{2-7}
\textbf{base}                              & 62.04            & 75.92                             & \textbf{45.46}      & 64.73            & 34.91            & 1                                     \\
\textbf{CSPDarknet}                        & \textbf{62.37 }  & \textbf{76.71}                    & 44.98               & 65.42            & 20.10            & 1.74                                  \\
\textbf{Darknet}                           & 60.80            & 76.28                             & 41.12               & 65.00            & 23.51            & 1.48                                  \\
\textbf{MobilenetV1}                       & 61.12            & 74.63                             & 45.30               & 63.42            & ~ 8.84           & 3.95                         \\
\textbf{MobilenetV2}                       & 60.35            & 74.96                             & 43.78               & 62.31            & ~ 8.84           & 3.95                         \\
\textbf{ResNet}                            & 60.92            & 75.37                             & 42.94               & 64.45            & 12.28            & 2.84                                  \\
\textbf{ResNeXt}                           & 59.96            & 75.53                             & 42.40               & 61.95            & 12.30            & 2.84                                  \\
\textbf{ShufflenetV1}                      & 59.74            & 74.84                             & 42.98               & 61.74            & ~ 9.78           & 3.57                                  \\
\textbf{ShufflenetV2}                      & 58.50            & 74.30                             & 39.15               & 62.06            & ~ \textbf{7.80 } & \textbf{4.48}                                  \\
\textbf{SqueezeNext}                       & 52.19            & 66.81                             & 39.34               & 50.43            & 14.66            & 2.38                                  \\
\textbf{Xception}                          & 61.12            & 76.13                             & 41.56               & \textbf{65.66}   & 10.72            & 3.26                                  \\
\bottomrule
\end{tabular}
\end{table}

\begin{figure*}[t!]
\centering
\includegraphics[width=0.75\textwidth]{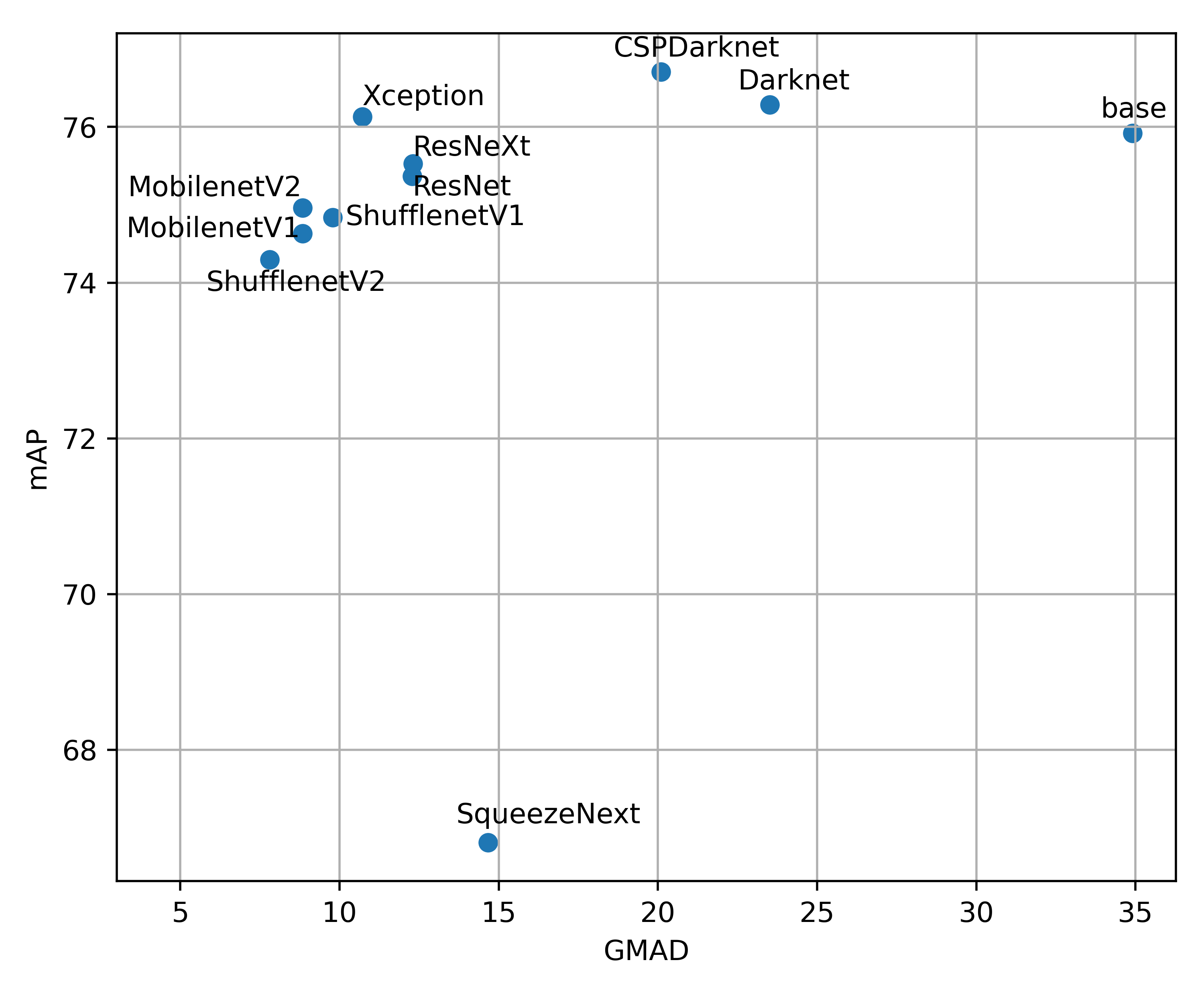}
\caption{mAP results (for car class) compared to the number of multiply-add operations for various backbone types. GMAD unit denotes $10^{9}$ of multiply-add operations. The Pareto set includes ShufflenetV2, MobilenetV2, Xception and CSPDarknet.}
\label{fig:map_vs_mad_car}
\end{figure*}

Figure \ref{fig:map_vs_mad_overall} shows the dependence of mAP on the number of multiply-add operations for each backbone type.
If we consider a multi-criteria optimization task where we maximise mAP and minimise the number of multiply-add operations then in the Pareto set \footnote{Pareto set is the set of non-dominated solutions} the following backbone types are found: CSPDarknet, MobilenetV1, ShufflenetV2.
Interestingly, the original version of PointPillars is not in the Pareto set because it is dominated by CSPDarknet.


\begin{figure*}[t!]
\centering
\includegraphics[width=0.75\textwidth]{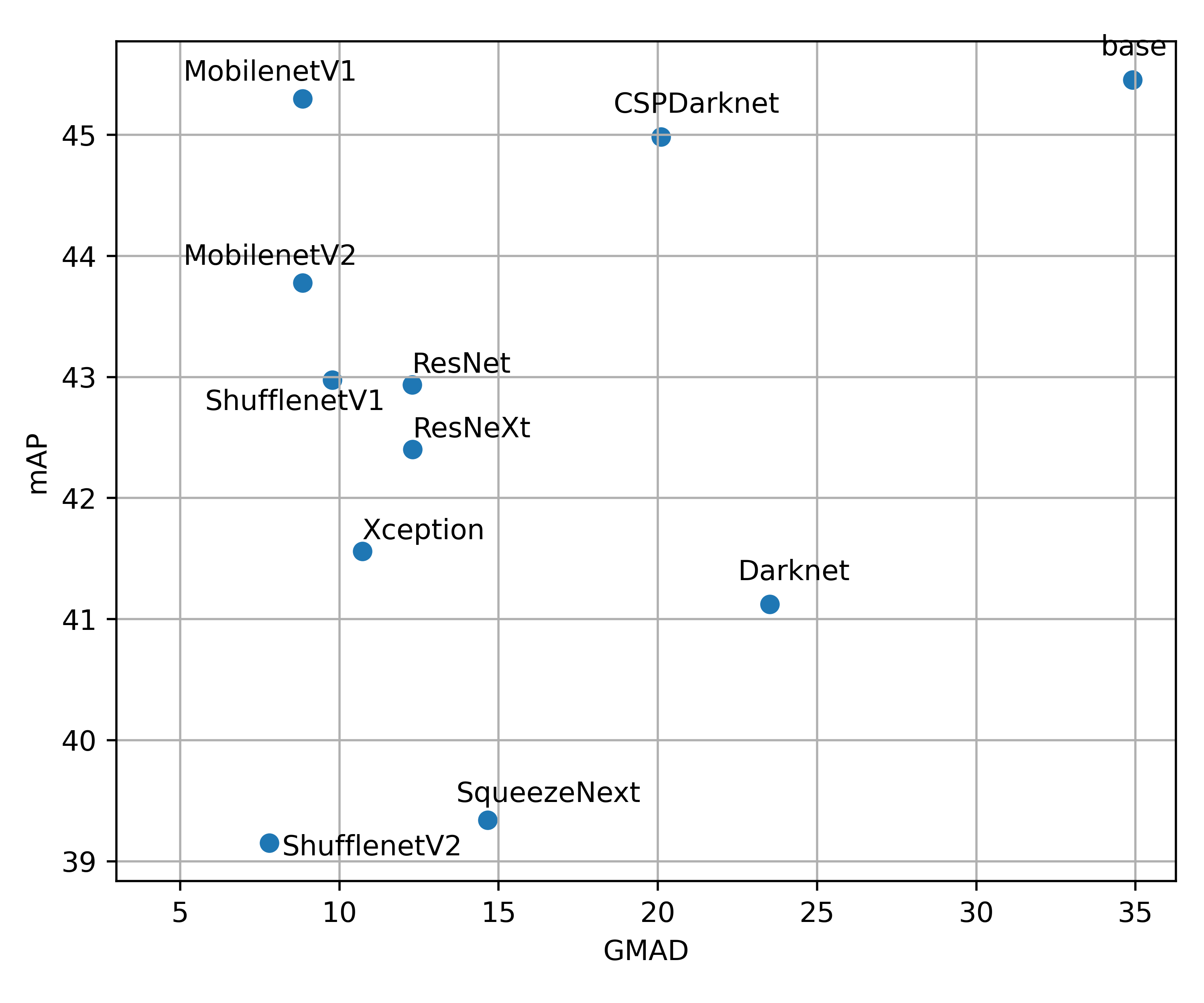}
\caption{mAP results (for pedestrian class) compared to the number of multiply-add operations for various backbone types. GMAD unit denotes $10^{9}$ of multiply-add operations. The Pareto set includes ShufflenetV2, MobilenetV1 and the base backbone.}
\label{fig:map_vs_mad_pedestrian}
\end{figure*}

The fastest backbone type is ShufflenetV2 and the most accurate is CSPDarknet; MobilenetV1 has intermediate values.
The answer to the question which one is best for implementation in an embedded system is not a~clear-cut and depends on the specific situation.
MobilenetV1 offers a very good compromise between detection efficiency and processing speed.
Its architecture is the simplest of those considered and is suitable for implementation on virtually any embedded platform and accelerator, including pipeline accelerators on FPGAs.
Compared to the original PointPillars version, MobilenetV1 is nearly 4x faster with a mAP decrease of only 1.13\%.
However, if we do not care so much about performance and more about speed, or we care more about detection efficiency and less about speed, then ShufflenetV2 or CSPDarknet, respectively, may be a better choice.
Compared to the original PointPillars version, ShufflenetV2 is nearly 4.5x faster with a 3.54\% decrease in mAP and CSPDarknet is more than 1.5x faster with a 0.33\% increase in mAP.


\begin{figure*}[t!]
\centering
\includegraphics[width=0.75\textwidth]{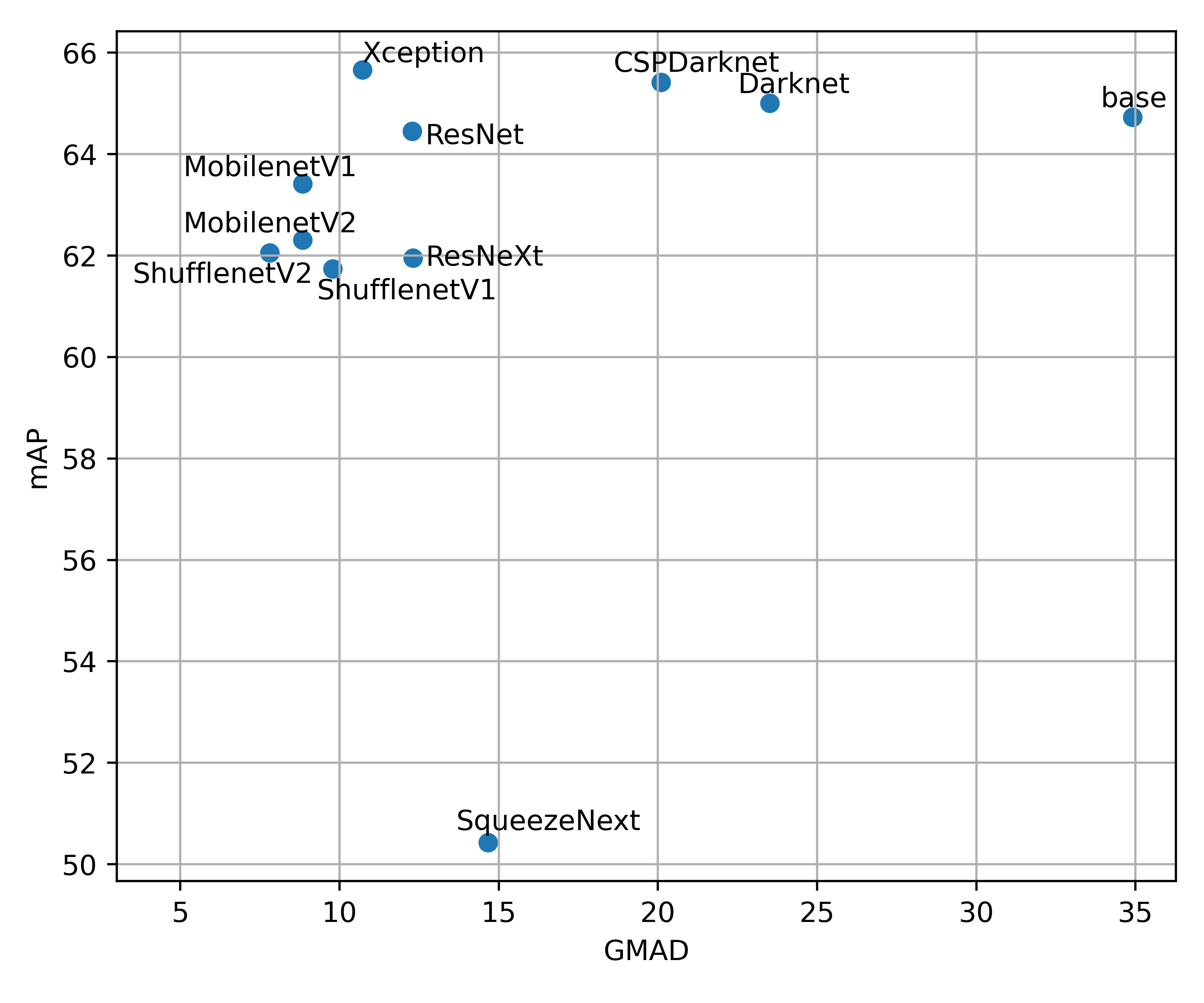}
\caption{mAP results (for cyclist class) compared to the number of multiply-add operations for various backbone types. GMAD unit denotes $10^{9}$ of multiply-add operations. The Pareto set includes ShufflenetV2, MobilenetV1 and Xception.}
\label{fig:map_vs_mad_cyclist}
\end{figure*}

If we want to detect objects of a particular class, the conclusions are slightly different from those drawn above.
For particular classes, the Pareto set of the corresponding multi-criteria optimisation problem (where the mAP metric applies to only one class) are:
\begin{itemize}
    \item for car class (see Figure \ref{fig:map_vs_mad_car}): 
    \begin{itemize}
        \item least multiply-add -- ShufflenetV2,
        \item intermediate values -- MobilenetV2, Xception,
        \item highest mAP -- CSPDarknet,
    \end{itemize}
    \item for pedestrian class (see Figure \ref{fig:map_vs_mad_pedestrian}):
    \begin{itemize}
        \item least multiply-add -- ShufflenetV2,
        \item intermediate values -- MobilenetV1,
        \item highest mAP -- base,
    \end{itemize}
    \item for cyclist class (see Figure \ref{fig:map_vs_mad_cyclist}):
    \begin{itemize}
        \item least multiply-add -- ShufflenetV2,
        \item intermediate values -- MobilenetV1,
        \item highest mAP -- Xception,
    \end{itemize}
\end{itemize}

ShufflenetV2 is always the fastest solution as, regardless of the class we are considering, the architecture of the individual solutions does not change.
However, mAP changes, and depending on the class we focus on, there are other types of backbones in the Pareto set, from which we should choose a solution for our particular application.

If a particular solution imposes memory constraints, one should also consider the number of parameters of a model (see Table \ref{tab:results_speed}).
Depending on backbone type, the checkpoint size varies from 3.7MB (ShufflenetV2, least number of parameters) to 19.4MB (base, the most number of parameters).
The model size can be further reduced by pruning (removing near-zero weights), quantisation (precision lower than default 32 bit floating point number) or compression.
However, even the highest obtained model size, 19.4MB, should not exceed memory limits in most of the solutions.



\begin{figure}[!t]
    \centering
    \includegraphics[width=0.75\textwidth]{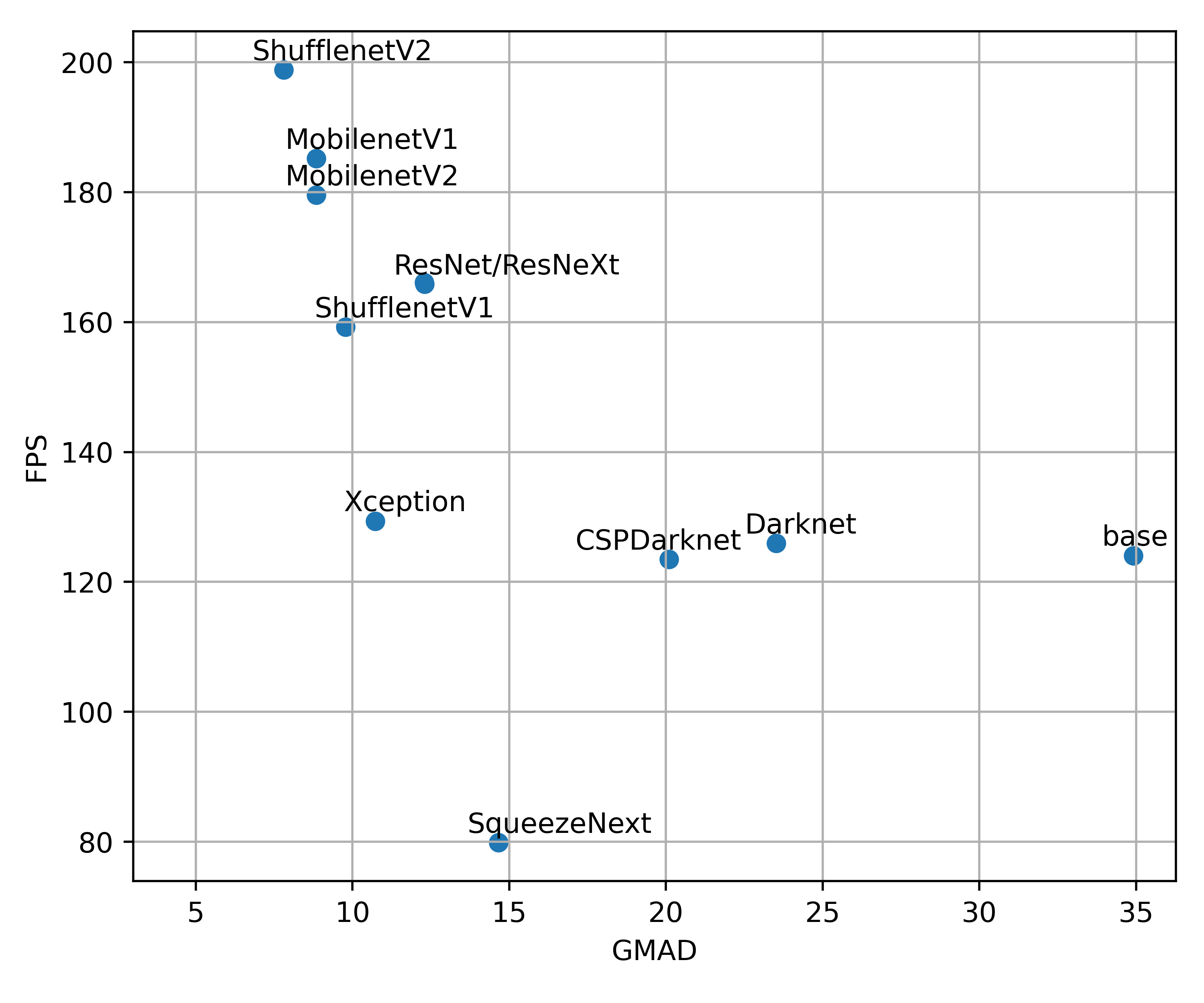}
    \caption{Frames per second vs number of multiply-add operations for different types of PointPillars backbone. GMAD unit denotes $10^{9}$ of multiply-add operations. The fps value takes into consideration only backbone, time of other PointPillars parts is not included. The general trend is broken probably due to specifics of the computing libraries and the computing platform.}
    \label{fig:fps}
\end{figure}

In Figure \ref{fig:fps}, we compare the number of backbone fps with the number of multiply-add operations for the backbones of the different PointPillars versions (for detailed data see Table \ref{tab:results_speed}).
The results are obtained with \emph{mmdetection3d} framework, the timing was measured on an Nvidia RTX 2070S GPU.
Only the backbone duration was included in the fps count, analysis of the whole algorithm fps results is described below.
As you can see, in this implementation, not always a lower multiply-add count means a faster network.
This is probably due to the specifics of the computing libraries and the computing platform -- some architectures are a better fit, others a worse one.
On other embedded platforms, such as FPGAs, the dependence of fps on the number of multiply-add may look quite different.

The speedup in terms of backbone fps (see Table \ref{tab:results_speed}) translates into only a slight acceleration in terms of the whole algorithm fps.
E.g., in case of MobilenetV2, 1.5x backbone speedup results in only a 1.14x whole algorithm acceleration.
This is related to the share of the backbone in the total processing time.
In the \emph{mmdetection3d's} implementation, the backbone of the ``base'' version is responsible for 39\% of the whole processing time.
According to the Amdahl's law \cite{prawo_amdahla}, optimising the backbone speed alone we can achieve up to 1.61x speedup of the whole algorithm (when backbone time approaches zero).
Regarding the fact that backbone includes 84\% of all multiply-add operations, 39\% is a small fraction.
Other most time consuming parts of the algorithm are Pillar Feature Net -- 10\%, the upsampling part after the ``top-down'' part of the backbone (in \textit{mmdetection3d} called a \textit{neck}) -- 15\% and NMS -- 25\%.
High processing times of the PFN and the neck are probably related to the specifics of the computing libraries and the computing platform.
In case of NMS, high time complexity is caused by it's sequential nature.
In this implementation, the processing time is scattered to many parts of the algorithm -- in order to achieve a large total speedup, one should accelerate all of them.

In our previous FPGA implementation \cite{stanisz_2021}, the backbone takes ca. 70\% time of the algorithm.
It allows us to achieve up to 3.33x speedup -- according to the Amdahl's law.
Accelerating backbone alone is still not enough for real-time, as theoretical 3.33x speedup of this algorithm would result in 8.89 fps.
On the other hand, accelerating other parts of the algorithm allows us to achieve up to 1.43x speedup, that translates into up to 3.82 fps for the algorithm.
Thus, one has to accelerate both backbone and other parts of the algorithm to achieve the desired 10 fps for real-time processing.

One should also keep in mind that after quantisation of examined PointPillars versions, mAP may change and some other model may become optimal in mAP sense.
Therefore, when quantisation is necessary (e.g. to deploy model on FPGA or speed up inference on GPU or eGPU), one should choose a subset of models suitable for a given hardware platform and reevaluate mAP afterwards.


\section{Conclusions}
\label{sec:conclusions}

In this paper, we have presented experiments on changing the backbone in a LiDAR-based 3D object detector to speed up the computations.
We investigated 10 versions of the PointPillars network, where each architecture was inspired by solutions from the computer vision domain (object detection and classification).
We used \emph{PyTorch} and the \emph{mmdetection3d} framework.


The results obtained indicate that it is possible to significantly speed up PointPillars with only a small decrease in accuracy (mAP value).
The version with the MobilenetV1-like backbone runs almost 4x faster than the original with a mAP decrease of 1.13\%.
Furthermore, the version with the CSPDarknet backbone runs more than 1.5x faster with an increase in mAP of 0.33\%.
Finally, an almost 4.5x speed-up with a mAP decrease of 3.54\% was achieved with a ShufflenetV2-like backbone.


The results presented above cover all classes of the KITTI dataset (car, pedestrian, cyclist).
However, while considering only one or selected classes, the changes in mAP may vary.
The results are also likely to be different with other datasets, such as nuScenes or Waymo.
However, this research indicates that it is worth at least considering using a ``lighter'' backbone architecture.
In the case of PointPillars, this increases the potential to implement the algorithm in a real-time embedded system while maintaining reasonable high detection performance.
Our previous work \cite{stanisz_2021} shows that such an implementation for the original version of the algorithm is very difficult if not impossible.

Compared to \cite{r12_fast_3d} and \cite{r13_realtime}, all of our considered PointPillars versions have lower processing time on a comparable GPU.
Given that \cite{r12_fast_3d} and \cite{r13_realtime} have smaller number of frames per second than the original PointPillars, it would be potentially harder to implement them on embedded devices in real-time.
On the other hand, work \cite{r11_pp_pruning} is focused on PointPillars pruning.
We first focused on changing the backbone to a lighter one, so that in the future we can perform pruning on a model with fewer operations than the original.
Potentially, this will allow us to obtain a higher algorithm speed than in \cite{r11_pp_pruning}, with minimal decrease in mAP.


One should keep in mind that speedup in sense of number of multiply-add operations does not directly translates into time speedup of the backbone and of the whole algorithm.
The resulting backbone's processing time is dependent on the hardware platform and implementation details, e.g. specifics of used libraries.
On the other hand, the whole algorithm's speedup is conditional on the backbone's processing time reduction and an initial share of the backbone in the whole processing time -- according to the Amdahl's law.


As part of our future work we would like, based on the best of the architectures considered, to use Network Architecture Search techniques to find even better (faster and more accurate) versions of the PointPillars network.
Afterwards, we will quantise optimised versions to enable implementation on FPGA devices and speed up inference on GPU/eGPU.
In order to reduce the number of multiply-add operations even further, we would like to use structural pruning, which potentially removes unnecessary computations from a particular trained network model.
The next step will be to optimise other parts of PointPillars, such as the PFN or NMS, to maximise the speed of the algorithm.
We also consider to enhance PointPillars, e.g. by using it as part of CenterPoint and implement it on an embedded system, e.g. on an SoC FPGA as a continuation of our previous work \cite{stanisz_2021}.

\subsubsection{Acknowledgements} The work presented in this paper was supported by the AGH University of Science and Technology project no. 16.16.120.773.

%
%
\bibliographystyle{bibtex/splncs03_fixed}
\bibliography{bibliography}

\end{document}